\documentclass[runningheads]{llncs}
\usepackage[T1]{fontenc}
\usepackage{graphicx}
\usepackage{booktabs}
\usepackage{subcaption}

\usepackage[acronym]{glossaries}
\newacronym{automl}{AutoML}{Automated Machine Learning}
\newacronym{dcai}{DCAI}{Data Centric Artificial Intelligence}
\newacronym{greenautoml}{Green AutoML}{Green Automated Machine Learning}
\newacronym{greenai}{Green AI}{Green Artificial Intelligence}
\newacronym{ml}{ML}{Machine Learning}
\newacronym{ea}{EA}{Evolutionary Algorithm}
\newacronym{edca}{EDCA}{Evolutionary Data Centric AutoML}
\newacronym{dr}{DR}{Data Reduction}
\newacronym{fs}{FS}{Feature Selection}
\newacronym{is}{IS}{Instance Selection}
\newacronym{hpo}{HPO}{Hyperparamenter Optimisation}
\newacronym{cash}{CASH}{Combined Algorithm Selection and Hyper-parameter Optimisation}
\newacronym{ai}{AI}{Artificial Intelligence}
\newacronym{ga}{GA}{Genetic Algorithm}

\usepackage{hyperref}
\usepackage{color}

\begin{document}
\title{EDCA -- An Evolutionary Data-Centric AutoML Framework for Efficient Pipelines}
\titlerunning{EDCA - A Framework for Efficient Pipelines}
%

\author{Joana Simões\orcidID{0000-0002-6846-3517} \and
João Correia\orcidID{0000-0001-5562-1996}}
\authorrunning{J. Simões and J. Correia}
%
\institute{University of Coimbra, CISUC/LASI – Centre for Informatics and Systems of the University of Coimbra, Department of Informatics Engineering\\
\email{{\{joanasimoes,jncor\}}@dei.uc.pt}
}
\maketitle              
\begin{abstract}

Automated Machine Learning (AutoML) gained popularity due to the increased demand for Machine Learning (ML) specialists, allowing them to apply ML techniques effortlessly and quickly. AutoML implementations use optimisation methods to identify the most effective ML solution for a given dataset, aiming to improve one or more predefined metrics. However, most implementations focus on model selection and hyperparameter tuning. Despite being an important factor in obtaining high-performance ML systems, data quality is usually an overlooked part of AutoML and continues to be a manual and time-consuming task. This work presents EDCA, an Evolutionary Data Centric AutoML framework. In addition to the traditional tasks such as selecting the best models and hyperparameters, EDCA enhances the given data by optimising data processing tasks such as data reduction and cleaning according to the problems' needs. All these steps create an ML pipeline that is optimised by an evolutionary algorithm.
To assess its effectiveness, EDCA was compared to FLAML and TPOT, two frameworks at the top of the AutoML benchmarks. The frameworks were evaluated in the same conditions using datasets from AMLB classification benchmarks. EDCA achieved statistically similar results in performance to FLAML and TPOT but used significantly less data to train the final solutions. 
Moreover, EDCA experimental results reveal that a good performance can be achieved using less data and efficient ML algorithm aspects that align with Green AutoML guidelines.

\end{abstract}

\glsresetall

\section{Introduction}
\label{sec:introduction}

\gls{ml} has grown exponentially in the last few years and has been used in various industries. The increased data availability pressures \gls{ml} experts to create systems to handle it in response to their needs. However, building these \gls{ml} systems is time-consuming and includes a lot of repetitive tasks. Hence, \gls{automl} appears in response to this high demand for experts, promising to reduce the time wasted on creating optimised \gls{ml} systems \cite{Truong_2019_Towards}. \gls{automl} frameworks use various optimisation algorithms to identify the most effective model or pipeline for the given \gls{ml} task \cite{Wang_2019_HumanAI}.  Their main focus has been in the modelling and hyper-parameter tuning phases \cite{Majidi_2022_An}, neglecting other essential pipeline steps \cite{Barbudo_2023_Eight}. 
Despite the benefit of automating the search for the best configuration, \gls{automl} has the downside of demanding a high computational power to evaluate all intermediate solutions \cite{Tornede_2021_Towards}.
Thus, \gls{greenautoml} appeared in response to the current concerns about the computational costs of \gls{automl}. It focuses on reducing the computational power required during the search and, at the same time, in ways to create more energy-efficient solutions.

Additionally, most \gls{automl} frameworks ignore the importance of removing data inconsistencies from the train data and do not optimise the step of the pipeline. Despite data processing is a time-consuming and manual task required before the \gls{automl} optimisation \cite{Xin_2021_Whither,Zoller_2021_Benchmark}. Normally, it is done manually or hard-coded on the frameworks where it is not customised to the data's requirements. 
\gls{dcai} proposes a shift from the \gls{automl}'s model-centric perspective into a data-centric perspective, based on the presumption that improving the data quality increases the accuracy of \gls{ml} systems \cite{Zha_2023_DataCentric,Bilal_2022_AutoPrep}. Another important data-centric task is to reduce data. Data selection helps to reduce data to only the most relevant one, without losing its information. At the same time, it also reduces the computational costs of training \gls{ml} systems.

This paper proposes \gls{edca}, a framework that combines the principles of \gls{automl}, \gls{dcai}, and \gls{greenautoml} to create a low-cost \gls{automl} framework capable of building efficient \gls{ml} pipelines for classification problems. It replicates the process made by \gls{ml} experts to create \gls{ml} pipelines. It starts by analysing the received data to define the data processing required to handle it. The processing steps required will be integrated into the pipeline structure. Additionally, the pipeline includes steps to select the most relevant data for the problem, both at the instances and features level. This data selection also helps reduce the computational costs during the optimisation. The pipeline ends with a classification model. With the created pipeline structure, an \gls{ea} is applied to find its best configuration.
Experiments with a subset of the AMLB benchmark datasets \cite{Gijsbers_2024_AMLB} indicate that \gls{edca} can achieve simple and efficient solutions for the same predictive performance that use significantly less data.

In this paper, we explored the importance of integrating \gls{dcai} techniques in a \gls{automl} framework. In addition, we studied ways to lower the computational costs involved with training based on \gls{greenautoml} guidelines. The contributions are the following: (i) proposal of an \gls{automl} framework, \gls{edca}, for the optimisation of a \gls{ml} pipeline using \gls{dcai} techniques to increase data quality and \gls{greenautoml} guidelines to reduce computational costs; (ii) experiments with the \gls{edca} and comparison with state-of-the-art \gls{automl} frameworks on benchmark datasets; (iii) analysis of the \gls{automl} frameworks regarding their predictive power, costs (amount of data use), number of solutions evaluated by each one and final solutions achieved; (iv) based on the analysis of experimental results, \gls{edca} promotes efficient solutions without losing predictive performance, enforcing the ideas of \gls{dcai} e \gls{greenautoml}.

The remainder of the paper is organised as follows. Section \ref{sec:related-work} describes the concepts of \gls{automl}, \gls{dcai}, and \gls{greenautoml}. Then, Section \ref{sec:edca} presents \gls{edca} in detail. The experimental study and the results achieved are presented in Section \ref{sec:experiments}. Lastly, Section \ref{sec:conclusion} describes the main conclusions and future work.

\section{Related Work}
\label{sec:related-work}

\gls{automl} appeared in response to the high demand of experts to perform \gls{ml} tasks \cite{He_2021_AutoML}. 
Its goal is to reduce the time wasted on repetitive tasks to find the best pipeline for a given \gls{ml} task \cite{Thornton_2013_AutoWEKA}. 
Several implementations are currently working to address \gls{automl}'s challenges. They differ on the parts of the \gls{ml} pipelines they optimised and on the algorithms used \cite{Barbudo_2023_Eight}. However, the search spaces increase with additional phases being optimised, making the problem harder \cite{Neutatz_2022_Data}.
Most frameworks specialise in the model selection and \gls{hpo} phases, with users handling the remaining steps manually, despite occasionally incorporating other phases into their process \cite{Majidi_2022_An,Zoller_2021_Benchmark}.

Arguably one of the first attempts of an \gls{automl} framework, Auto-Weka \cite{Thornton_2013_AutoWEKA} uses Bayesian optimisation to identify the optimal pipeline and applies meta-learning from past experiments to new datasets. 
Similarly, Auto-sklearn \cite{Feurer_2019_Autosklearn}, built on top of Scikit-learn, by including some data-processing steps on the search space to transform the input data and by creating an ensemble with the tested pipelines. However, in later versions, most of the data preprocessing was excluded \cite{Feurer_2022_AutoSklearn2,Gijsbers_2024_AMLB}.
AutoGluon uses a three-layer stacking ensemble with predefined models, not focusing on hyperparameter tuning \cite{Erickson_2020_AutoGluonTabular}. It assumes some models are better, and only trains theoretically best ones when time is limited. AutoGluon also conducts an initial dataset analysis for data processing, but these steps are fixed and not optimised.
TPOT, another known framework, represents the \gls{ml} pipeline in a graph structure and uses genetic programming (GP) \cite{Olson_2016_Evaluation}. TPOT allows dynamic pipelines created by data preprocessing and predictive operators. However, despite using mechanisms to control pipelines' complexity, the dynamic pipelines increase the search space because the optimisation must simultaneously identify the optimal structure and configuration of the operators. 
FEDOT also applies GP to optimise dynamic pipelines that could have more than one predictive model, using a multi-objective evaluation function to control complexity and increase performance. 
Likewise, AutoML-DSGE \cite{Assuno_2020_Evolution} optimises dynamic \gls{ml} pipelines through grammatical evolution and DeepLine \cite{Heffetz_2020_DeepLine} applies reinforcement learning to optimise similar dynamic pipelines.
Additionally, Wang et al. created FLAML \cite{Wang_2021_FLAML} to reduce the computational costs of evaluating several intermediate solutions. The framework focuses on selecting the ideal models, using the estimated cost (CPU time wasted) and performance in previous evaluations to make the decisions. It also uses an incremental instance reduction to lower the costs of the evaluations. 
More \gls{automl} exists and can be consulted in surveys such as \cite{Zoller_2021_Benchmark,He_2021_AutoML,Baratchi_2024_AutoML}.

\gls{automl} is often criticised for its expensive and resource-consuming search. The frameworks often evaluate similar intermediate solutions without significant improvements in performance.
\gls{greenautoml} introduces guidelines to create environmentally friendly \gls{automl} frameworks \cite{Tornede_2021_Towards}.
It identifies effective strategies to minimise resources and lower computational costs usually associated with \gls{automl}. One important guideline is to use less data to test intermediate solutions since their energy consumption and training time are positively correlated with the size of the training data used \cite{Verdecchia_2022_DataCentric}.
Therefore, reducing data is environmentally beneficial and accelerates the optimisation. 
Additional strategies refer to the use of multi-objective evaluation to incorporate costs on the decision process, and in ways to prevent and overcome performance stagnation \cite{Castellanos_2023_Improving,Tornede_2021_Towards}.

Data quality is essential \cite{Zha_2023_DataCentric} and it influences the quality of the \gls{ml} systems trained with it \cite{Li_2021_CleanML,Krishnan_2015_SampleClean}.
Thus, \gls{dcai} proposes shifting from the old model-centric approach to handle \gls{ml} systems into a more data-centric method, where the focus is improving data quality and, respectively, the  \gls{ml} systems.
Depending on the data's needs, various data processing tasks need to be applied such as imputing missing values, removing duplicates, encoding categorical features and normalising data \cite{Xin_2021_Whither}. 
Despite being time-consuming, few frameworks tried to automate these tasks due to the increase in the search space \cite{Xin_2021_Whither,Krishnan_2015_SampleClean,Li_2021_CleanML}.  
In most \gls{automl}, they are not optimised to the problem's needs, they are hard-coded or required to be manually handled \cite{Zoller_2021_Benchmark}. Despite recognising data importance, they minimise the problem by creating robust models that handle noisy data \cite{Whang_2023_DataCollection,Neutatz_2022_Data}.
Additionally, large datasets can sometimes have lower predictive performance due to increased noise and complexity \cite{Nalepa_2018_Genetic,Tong_2011_Determining,Hoens_2013_Imbalanced}. \gls{dr} mitigates this by selecting only relevant instances or features, preserving the original data properties while reducing the computational costs of training \gls{ml} systems \cite{Verdecchia_2022_DataCentric,Tornede_2021_Towards}. \gls{dr} techniques include \gls{is}, which narrows data to representative instances (samples), and \gls{fs}, which removes redundant features, improving the model's generalisation ability. \gls{ea} have been used for both \gls{is} \cite{Rathee_2019_Instance,Cano_2003_Using,Reeves_2001_Using,Garca_2009_Evolutionary,Triguero_2016_Evolutionary,Fernandes_2019_Evolutionary} and \gls{fs} \cite{Xue_2016_A,Huang_2010_MultiObjective,Tan_2014_MultiObjective,Souza_2011_CoEvolutionary}, helping to select optimal data subsets. Some works also combine \gls{is} and \gls{fs} \cite{Souza_2008_Novel,Derrac_2009_A} with\gls{ea}, but maintain separate populations that are only integrated at the evaluation stage. Additionally, some \gls{automl} frameworks integrate \gls{fs} into their pipeline, optimising \gls{fs} algorithms similarly to predictive models\cite{Olson_2016_Evaluation,Thornton_2013_AutoWEKA,Feurer_2019_Autosklearn}. When \gls{is} appears in \gls{automl} frameworks, it is usually a random \gls{is} to accelerate the optimisation \cite{Wang_2021_FLAML}.

To the best of our knowledge, most \gls{automl} frameworks ignore the importance of data and continue with a model-centric perspective. Nevertheless, optimising the data used by the \gls{automl} frameworks has benefits in performance and costs. \gls{dcai} showed how removing inconsistencies and transforming data into a more digestible format may increase performance. The necessary preprocessing tasks may vary depending on the problem and should be tailored to each. At the same time, \gls{dr} techniques aim to select only the best data to help increase performance and the generalisation ability. These \gls{dr} techniques are also recommended by the \gls{greenautoml} community, as they also allow to train more solutions and achieve final solutions with less computational costs associated. 

\section{EDCA -- Evolutionary Data Centric AutoML}
\label{sec:edca}
\begin{figure}[t!]
    \centering
    \includegraphics[width=1\linewidth]{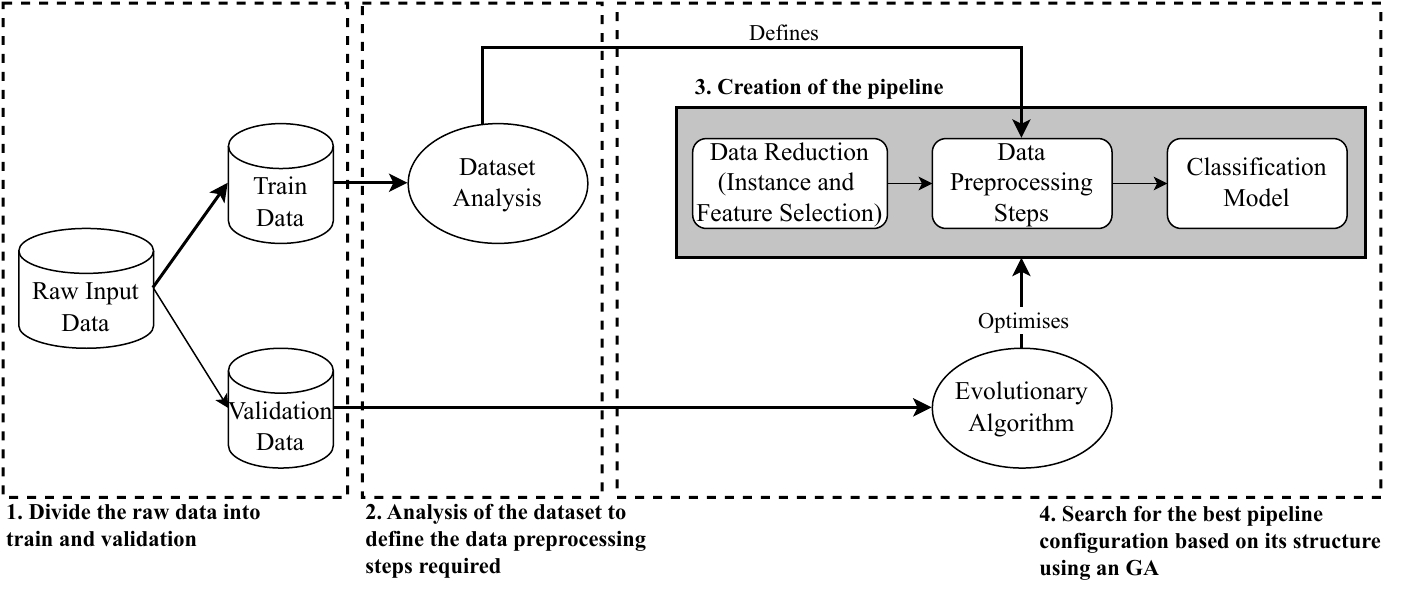}
    \caption{Process used by EDCA.}
    \label{fig:framework-overview}
\end{figure}

\gls{edca}\footnote{EDCA's \href{https://github.com/PugtgYosuky/EDCA}{GitHub} repository} is an \gls{automl} framework capable of creating a complete \gls{ml} pipeline, attending to the principles of \gls{dcai} \cite{Zha_2023_DataCentric} and \gls{greenautoml} \cite{Tornede_2021_Towards,Verdecchia_2022_DataCentric,Castellanos_2023_Improving}, resorting to an automated analysis module for the data combined with an \gls{ea} that evolves the complete pipeline, model and data. 
Figure \ref{fig:framework-overview} presents \gls{edca}'s process. The input data is first divided into train and validation (1). Then, the \gls{ml} pipeline steps are created based on an initial train data analysis (3), which determines the preprocessing required. Additionally, the pipeline includes \gls{dr} steps, which helps to select only the relevant data and reduce the optimisation costs (3). The pipeline structure ends with the optimisation process of a model (3). 
A \gls{ga} is used to find the best pipeline configuration for the problem within a set time limit (4). 

\gls{edca}'s process starts by dividing the input data into train and validation data (Phase 1). The train data passes to Phase 2, whereas the validation data is only used in Phase 4 to assess individuals' predictive power. 

In Phase 2, the data is analysed to assess which processing techniques are required for the given data. It determines the features, types, and characteristics of the data. 
Each feature in the data could belong to one of four groups: binary,  categorical, numerical or identifier. The data characteristics help to understand the types of preprocessing needed for the problem, and each group of features will have a different treatment.
After the analysis, a pipeline structure is created (Phase 3). In total, five different preprocessing steps could exist in the pipeline. If the analysis indicates that the data contains numerical features, a normalisation step will be added. The columns with identifiers will be removed since they do not add relevant information. When categorical features exist, an encoder step is added, as the models only digest numerical data. When there are features with missing data, several imputation steps are added to the pipeline, one for each feature type, since the hyperparameters will be different.

Creating the pipeline's structure based on the data characteristics allows for using only the minimum required preprocessing steps (Phase 3). If all the preprocessing steps were available every time, time would be spent tuning irrelevant methods for the data in question. Thus, using an initial dataset analysis to define the preprocessing required and corresponding search space helps reduce the problem's complexity and only spend time tuning the essential \gls{ml} steps. During the optimisation, these processing steps are always present in the individuals.

The \gls{ml} pipeline includes steps for \gls{dr} (Phase 3), which allows to select of only the most relevant data for the problem, while, at the same time, reducing the costs and complexity of the problem by accelerating the optimisation. \gls{edca} can include two \gls{dr} steps for the \gls{is} and \gls{fs}. However, since using them may or may not be beneficial, \gls{edca} uses automatic data optimisation.
The automatic data optimisation searches for the best combination of the \gls{dr} techniques by turning on or off these genes along the optimisation. It can be chosen to use only one type of \gls{dr}, both, or none of them. 
Phase 3 also adds a final step to the pipeline for a predictive model, similar to other frameworks. A suitable model should be used at this stage since it would impact the final results. 
Overall, the pipeline uses a sequential linear structure, with only one path for data to follow.

\begin{figure}[t!]
    \centering
    \includegraphics[width=\linewidth]{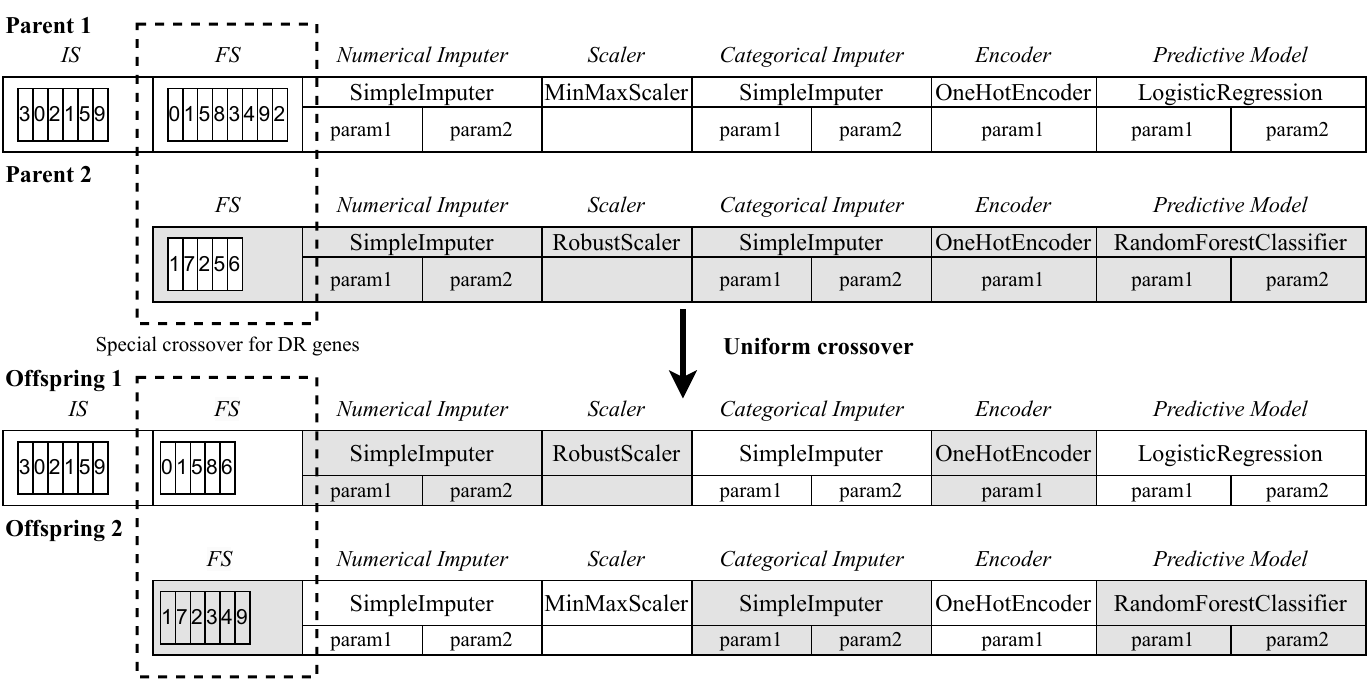}
    \caption{Example of crossover. It is not applied to the IS genes because they are absent in both individuals. For the FS genes, a gene-level one-point crossover is applied. The non-DR genes use a uniform crossover at the parent level.}
    \label{fig:crossover}
\end{figure}

To optimise the pipeline, \gls{edca} uses a \gls{ga} due to its effectiveness in large search spaces \cite{Tsai_2013_Genetic}. Each individual is represented by a list of genes and each gene represents a step of the \gls{ml} pipeline, including\gls{dr}, the different processing required, and a predictive model (see the example of Parent 1 in Figure \ref{fig:crossover} where it represents the maximum number of genes an individual could have).
The \gls{dr}-related genes are composed of a set of integer numbers corresponding to indices, since in previous works that used a binary representation for the \gls{dr} the reduction tented to reach only 50\% \cite{Rathee_2019_Instance}. These \gls{dr} genes may or not be present on the pipeline since we use a procedure called automatic data optimisation, that can turn on or off their \gls{dr} genes. When a \gls{dr} gene is present, it says which indices, related to an instance or feature, should be used to train the pipelines. The indices vary between zero and the maximum length possible, i.e.,  $[0, max_{instances}[$ or $[0, max_{features}[$.  The remaining genes correspond to the other pipeline steps and are composed of the model and its hyperparameters, where each model may vary on the number of hyperparameters. The individuals use a dynamic structure since they may vary depending on the presence or absence of the \gls{dr} genes, while the others are always present. \gls{edca} implements both crossover and mutation as variation operators.

The crossover operator is applied to interchange genes between individuals. It receives two parents and returns two offspring. The genes of the received individuals are switched between them using uniform crossover, except for the \gls{dr}-related genes (see Figure \ref{fig:crossover}). The \gls{dr} genes include a special crossover operator since they use a different representation and they might or may not be present in the individuals. 
The crossover is applied only if both individuals have the gene. In that case, a one-point crossover occurs at the gene level. If not, the gene is directly inherited from the original individual to the corresponding offspring. The one-point crossover preserves unique values to prevent data augmentation.

\begin{figure}[t!]
    \centering
    \includegraphics[width=\linewidth]{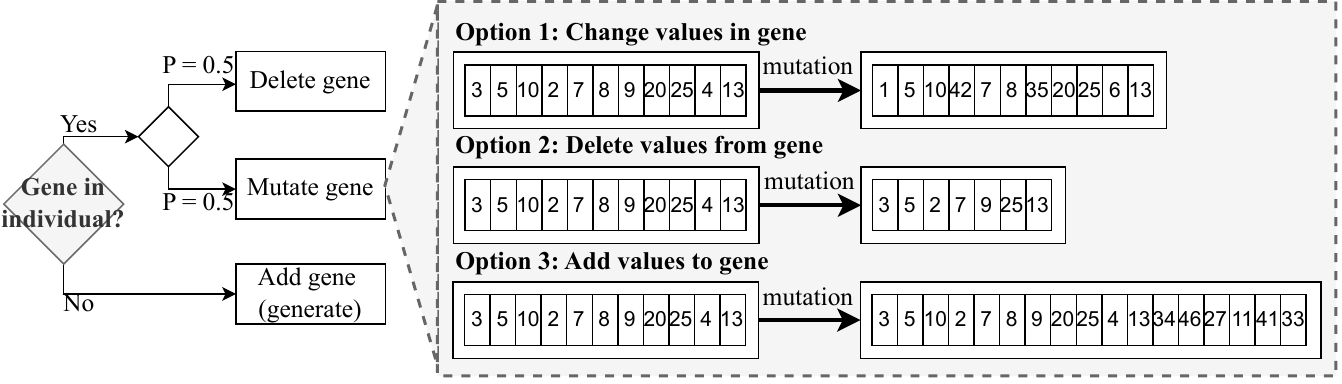}
    \caption{Flow chart of the mutation applied to \gls{dr} genes (IS or FS).}
    \label{fig:dr-mutation-automatic}
\end{figure}

Additionally, the mutation operator applied only changes one gene per individual. The general case works by changing the method in a step of the pipeline and its hyperparameters or changing only the hyperparameters, each option with the same likelihood. Since automatic data optimisation is applied to the \gls{dr} genes, depending on the presence or absence of the gene, different actions may be used (see Figure \ref{fig:dr-mutation-automatic}). If the gene is present, it can be deleted or changed. Otherwise, the only option is to add it. When it is chosen to mutate the gene, three different operations can be applied: (i) change $\theta$\% of the indices for other values; (ii) add $\theta$\% of new indices; (iii) delete $\theta$\% of the indices. The user defines the maximum percentage of change from the initial data size, and $\theta$\% varies between zero and the maximum defined in the runtime. 

Each individual's fitness is evaluated using the performance metric error, calculated with the validation data from Phase 1 validation in Figure \ref{fig:framework-overview}, similar to other frameworks. The fitness ranges from e to one ($[0, 1]$), where zero indicates the best performance and one the lowest.

Furthermore, the optimisation uses some techniques to save time and help the search process. 
To maintain diversity and avoid performance stagnation as recommended by the \gls{greenautoml} guidelines \cite{Tornede_2021_Towards}, the optimisation uses a patience mechanism to wait for improvements on the best solutions, where the population is restarted and new individuals substitute the current population when the performance of the best individuals does not improve after a certain number of generations. The \gls{ga} also uses elitism to save the best solutions across generations, even when a restart occurs.
Similar to other \gls{automl} frameworks, \gls{edca} offers parallel searching to evaluate multiple solutions simultaneously, which allows to accelerate the process and test more generations.

In short, \gls{edca} is composed of different parts (see Figure \ref{fig:framework-overview}). The entire pipeline consists of an initial \gls{dr} phase, followed by the preprocessing steps dependent on the analysis. Then, the pipeline ends with a predictive model. The pipeline structure is optimised by an \gls{ga} that explores different configurations during a specified time budget and attempts to optimise a performance metric.

\section{Experimental Study}
\label{sec:experiments}

An experimental study was carried out to compare \gls{edca} with state-of-the-art \gls{automl} frameworks, FLAML \cite{Wang_2021_FLAML} and TPOT \cite{Olson_2016_Evaluation}. This section details the setup used (Sub-section \ref{sec:experimental-setup}) and the results achieved (Sub-section \ref{sec:experimental-results}).

\subsection{Experimental Setup}
\label{sec:experimental-setup}

All the frameworks were evaluated under the same conditions, i.e., they had the same time budget and used the same optimisation metric for the search.
The Matthews Correlation Coefficient (MCC), due to being robust to the class used as positive \cite{Chicco_2020_Advantages,Jurman_2012_Comparison}, was selected to measure the performance during the optimisation and to assess frameworks' predictive ability. Table \ref{tab:common-parameters} and \ref{tab:evolutionary-parameters} present the parameters used for the implementations. \gls{edca}'s parameters were defined after evaluating several combinations using train-test divisions in different datasets. Any parameters not listed in Table \ref{tab:common-parameters} are set to default for FLAML and TPOT. 

\gls{edca} is currently focused on classification problems (binary or multi-class) with tabular data. 
Therefore, different classification models were considered at this step, with all having different hyperparameters to be tuned. The three frameworks used the same predictive models list, with small variations in the range of values of the hyperparameters. The implementations are from Scikit-Learn \cite{Pedregosa_2011_Scikitlearn}, \href{https://lightgbm.readthedocs.io/}{LightGBM} and \href{https://xgboost.readthedocs.io/}{XGBoost}. 

We used a subset of five classification datasets from Gijsbers et al. benchmark study of \gls{automl} frameworks \cite{Gijsbers_2024_AMLB}.
This subset was selected because the datasets required some data preprocessing work, and for that, each preprocessing type can be applied at least once (see Table \ref{tab:datasets-summary} with their characteristics).

\begin{table}[t]
    \centering
    \caption{Parameterisation used by the frameworks. FLAML does not use FS, and TPOT does not use IS since they do not implement them.}
    \begin{minipage}[t]{0.45\textwidth}
        \subcaption{Common parameters}
        \label{tab:common-parameters}
        \resizebox{\textwidth}{!}{
        \begin{tabular}{r|r}
    \toprule
         \textbf{Parameter} & \textbf{Value} \\
    \midrule
         Runs & 30 \\
         External Data Division & CV \\
         K-fold & 5 \\
         Stop criterion & Time Budget \\
         Time Budget (seconds) & 900\\
         Optimisation Metric & MCC \cite{Chicco_2020_Advantages,Jurman_2012_Comparison}\\
         Instance Selection & True \\
         Feature Selection & True \\
         Parallel Jobs & 5 \\
         Internal Data Division & Train-Val\\
         Validation Percentage & 25\% \\
    \bottomrule
    \end{tabular}
    }
    \end{minipage}
    \begin{minipage}[t]{0.45\textwidth}
    \subcaption{Evolutionary parameters}
    \label{tab:evolutionary-parameters}
    \resizebox{\textwidth}{!}{
    \begin{tabular}{r|r}
    \toprule
         \textbf{Parameter} & \textbf{Value} \\
    \midrule
         Probability of Mutation & 0.3\\
         Probability of Crossover & 0.7\\
         Elitism Size & 1\\
         Population Size & 50 \\
         Tournament Size & 3 \\
         Patience (no. generations) & 5 \\
         Automatic Data Optimisation & True \\
         Percentage of change & 10\% \\
    \bottomrule
    \end{tabular}
    }
    \end{minipage}
\end{table}

\begin{table}[t]
    \centering
    \caption{Summary of the datasets. Acronyms: ID: OpenML\cite{Vanschoren_2014_OpenML} ID; MV: Missing values; NF: Numerical features; CF: Categorical features; BF: Binary features.}
    \begin{tabular}{r|r|r|r|r|r|r|r|r}
    \toprule
         \textbf{Dataset} & \textbf{ID} &\textbf{Instances} & \textbf{Features} & \textbf{Classes}  & \textbf{MV} & \textbf{NF} & \textbf{CF} & \textbf{BF}\\
         \midrule
         Australian & 40981 & 690 & 14 & 2 & No & 14 & 0 & 0\\
         adult & 1590 &48842 & 14 & 2 & Yes & 6 & 7 & 1 \\
         cnae-9 & 1468 & 1080 & 856 & 9  & No & 856 & 0 & 0 \\
         credit-g & 31 & 1000 & 20 & 2 & No & 7 & 11 & 0 \\
         mfeat-factors & 12 & 2000 & 216 & 10 & No & 216 & 0 & 0\\
         \bottomrule
    \end{tabular}
    \label{tab:datasets-summary}
\end{table}

\subsection{Results}
\label{sec:experimental-results}

For each dataset, 30 independent runs were performed with different random seeds. A 5-fold cross-validation (CV) was applied to each run and the predictive performance in each fold was calculated.
To verify the statistical significance of the results obtained by \gls{edca} in comparison to the other frameworks, the Mann-Whitney statistical test was employed. A significance level of 99\% was used in the statistical tests, using a Bonferroni correction to reduce Type I errors.

The MCC values achieved by the frameworks are presented in Table \ref{tab:mcc-values}. There are almost no statistically significant differences in MCC values between FLAML and TPOT when compared to \gls{edca}. Additionally, TPOT fails in some datasets when used off-the-shelf without manual data processing.

\begin{table}[t!]
    
    \centering

    \caption{Average MCC values (± standard deviation) over 30 runs for EDCA, FLAML, and TPOT. Results for EDCA and FLAML after retraining on all data are included, while TPOT always uses all data. Statistically significant differences compared to EDCA are in bold. Arrows show changes after retraining.}

    \begin{tabular}{r|r|r|r|r|r}
    \toprule
    \textbf{Dataset} & \textbf{EDCA} & \textbf{FLAML} & \textbf{TPOT} & \textbf{EDCA-All} & \textbf{FLAML-All} \\
    \midrule
    Australian & 0.72±0.01 & 0.72±0.02 & 0.72±0.02 & \textbf{0.73±0.02}$\uparrow$ & 0.72±0.02 \\
    adult & 0.63±0.0 & 0.63±0.01 & - & \textbf{0.63±0.0} & 0.63±0.01 \\
    cnae-9 & 0.93±0.01 & 0.93±0.01 & \textbf{0.94±0.01} & \textbf{0.94±0.01}$\uparrow$ & 0.93±0.01 \\
    credit-g & 0.32±0.04 & 0.34±0.03 & - & \textbf{0.36±0.03}$\uparrow$ & 0.34±0.02 \\
    mfeat-factors & 0.97±0.0 & \textbf{0.96±0.0} & 0.97±0.0 & \textbf{0.97±0.0} & 0.96±0.0 \\
    \bottomrule
    \end{tabular}
    
    \label{tab:mcc-values}
\end{table}

Considering the percentage of data used in the end, \gls{edca} uses significantly less data than the remaining frameworks regarding total data, instances, and features in almost all the experiments (see Tables \ref{tab:percentage-data}, \ref{tab:percentage-samples} and \ref{tab:percentage-features}). This occurs since FLAML only uses incremental sampling to accelerate the optimisation, ignoring \gls{fs}. With this incremental sampling, it always reaches the maximum data available, i.e., the internal training data available, since 25\% is for validation and does not retrain with it. On the other hand, TPOT does not use \gls{is} and only applies \gls{fs}. 
Considering that \gls{edca} can achieve similar results to FLAML and TPOT with less data, in the context of \gls{greenautoml}, it is beneficial to use \gls{edca} due to the lower associated costs for identical predictive performance.

\begin{table}[t!]
    \centering
    \caption{Average percentage of data (± standard deviation) over 30 runs used by the frameworks. It includes the percentage of total, instances and features that were calculated by dividing the final data size by the original size. 
    Statistically significant differences compared to EDCA are highlighted in bold.}
    \begin{minipage}[t]{0.49\textwidth}
    \subcaption{Percentage of data}
    \label{tab:percentage-data}
    \resizebox{\textwidth}{!}{
        \begin{tabular}{r|r|r|r}
        \toprule
        \textbf{Dataset} & \textbf{EDCA} & \textbf{FLAML} & \textbf{TPOT} \\
        \midrule
        Australian & 0.35±0.06 & \textbf{0.75±0.0} & \textbf{1.02±0.05} \\
        adult & 0.64±0.04 & \textbf{0.75±0.0} & - \\
        cnae-9 & 0.59±0.05 & \textbf{0.75±0.0} & \textbf{0.98±0.06} \\
        credit-g & 0.4±0.09 & \textbf{0.75±0.0} & - \\
        mfeat-factors & 0.47±0.06 & \textbf{0.75±0.0} & \textbf{0.99±0.03} \\
        \bottomrule
        \end{tabular}
    }
    \end{minipage}
    \begin{minipage}[t]{0.49\textwidth}
        \subcaption{Percentage of instances}
        \label{tab:percentage-samples}
        \resizebox{\textwidth}{!}{
        \begin{tabular}{r|r|r|r}
        \toprule
        \textbf{Dataset} & \textbf{EDCA} & \textbf{FLAML} & \textbf{TPOT} \\
        \midrule
        Australian & 0.43±0.09 & \textbf{0.75±0.0} & \textbf{1.0±0.0} \\
        adult & 0.66±0.05 & \textbf{0.75±0.0} & - \\
        cnae-9 & 0.6±0.06 & \textbf{0.75±0.0} & \textbf{1.0±0.0} \\
        credit-g & 0.47±0.09 & \textbf{0.75±0.0} & - \\
        mfeat-factors & 0.67±0.05 & \textbf{0.75±0.0} & \textbf{1.0±0.0} \\
        \bottomrule
        \end{tabular}
        }
    \end{minipage}

    \begin{minipage}[b]{0.52\textwidth}
        \subcaption{Percentage of features}
        \label{tab:percentage-features}
        \resizebox{\textwidth}{!}{
        \begin{tabular}{r|r|r|r}
        \toprule
        \textbf{Dataset} & \textbf{EDCA} & \textbf{FLAML} & \textbf{TPOT} \\
        \midrule
        Australian & 0.84±0.1 & \textbf{1.0±0.0} & \textbf{1.02±0.05} \\
        adult & 0.98±0.03 & \textbf{1.0±0.0} & - \\
        cnae-9 & 0.98±0.03 & \textbf{1.0±0.0} & 0.98±0.06 \\
        credit-g & 0.86±0.11 & \textbf{1.0±0.0} & - \\
        mfeat-factors & 0.72±0.1 & \textbf{1.0±0.0} & \textbf{0.99±0.03} \\
        \bottomrule
        \end{tabular}
        }
    \end{minipage}
    \label{tab:percentages}
\end{table}

\begin{table}[t!]
    \centering
    \caption{Average MCC values (± standard deviation) for EDCA, FLAML, TPOT, and FLAML + EDCA and TPOT + EDCA (i.e., FLAML or TPOT retrained with EDCA-selected data). Statistically significant differences from EDCA results are in bold and the "*" reveals statistically significant differences from original framework's MCC values. Arrows show changes after retraining.}

    \begin{tabular}{r|r|r|r|r|r}
    \toprule
    \textbf{Dataset} & \textbf{EDCA} & \textbf{FLAML} & \textbf{TPOT} & \textbf{FLAML+EDCA} & \textbf{TPOT+EDCA} \\
    \midrule
    Australian & 0.72±0.01 & 0.72±0.02 & 0.72±0.02 & \textbf{0.69±0.04}*$\downarrow$ & \textbf{0.64±0.1}*$\downarrow$ \\
    adult & 0.63±0.0 & 0.63±0.01 & - & \textbf{0.63±0.0} & - \\
    cnae-9 & 0.93±0.01 & 0.93±0.01 & \textbf{0.94±0.01} & \textbf{0.91±0.01}*$\downarrow$ & \textbf{0.92±0.03}*$\downarrow$ \\
    credit-g & 0.32±0.04 & 0.34±0.03 & - & \textbf{0.24±0.07}*$\downarrow$ & - \\
    mfeat-factors & 0.97±0.0 & \textbf{0.96±0.0} & 0.97±0.0 & \textbf{0.95±0.0}*$\downarrow$ & \textbf{0.96±0.01}*$\downarrow$ \\
    \bottomrule
    \end{tabular}
    \label{tab:mcc-values-retrained}
\end{table}

\begin{figure}[t!]
    \centering
    \includegraphics[width=1\linewidth]{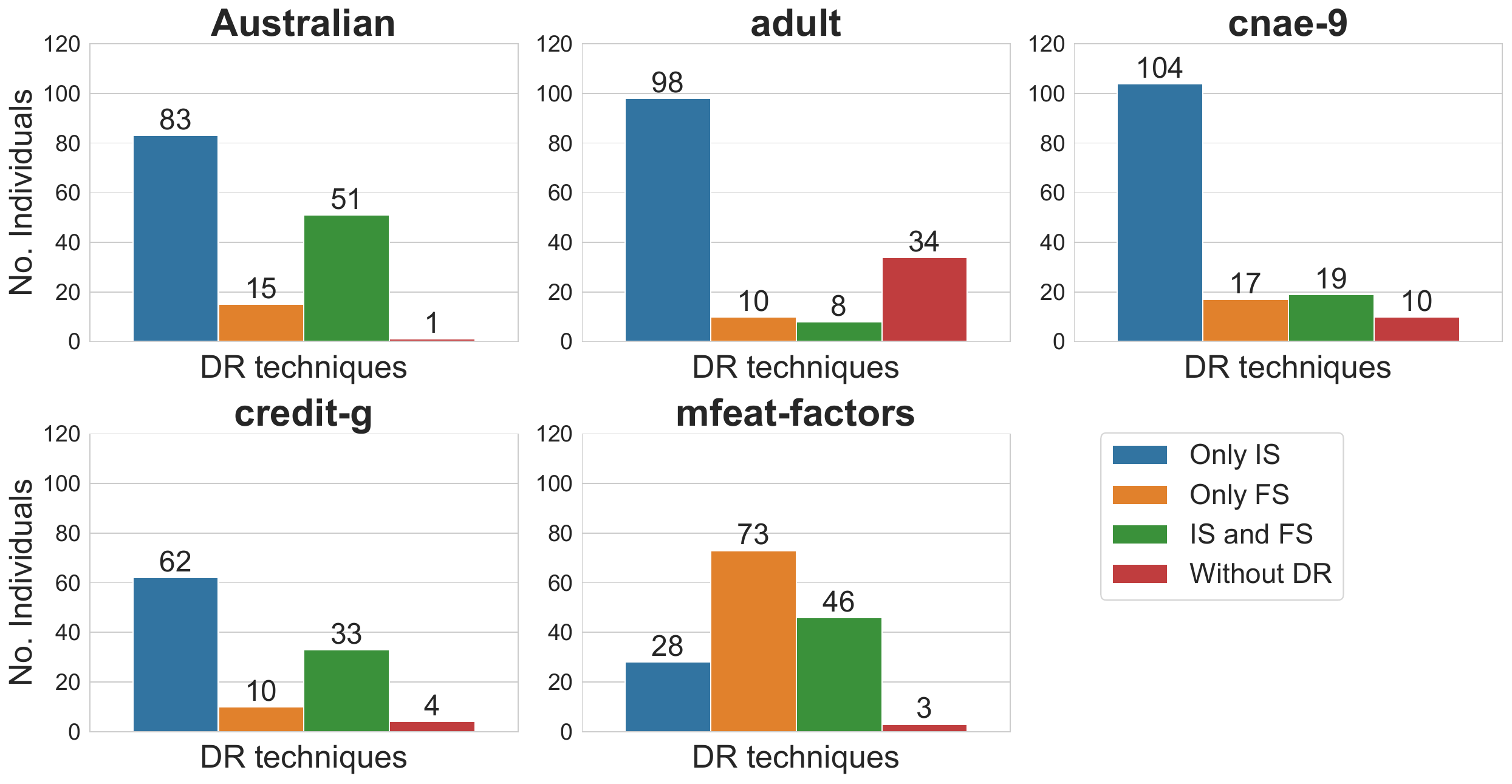}
    \caption{Distribution of EDCA's DR techniques in a 5-fold CV across 30 runs.}
    \label{fig:histogram-dr-techniques}
\end{figure}

Another relevant finding in \gls{edca} was that when comparing the percentage of instances (Table \ref{tab:percentage-samples}) and features (Table \ref{tab:percentage-features}) achieved, it is shown that the percentage of features is always higher than the percentage of instances, meaning that the features may be more relevant for the final results than the number of instances used. Furthermore, Figure \ref{fig:histogram-dr-techniques} shows the histogram of the \gls{dr} techniques applied on the final pipelines for each dataset, since they may vary because of the use of automatic data optimisation (see Section \ref{sec:edca}).
The results indicate that \gls{edca} accomplishes very different final solutions, which may related to the fact that it uses the best data available without any constraints. An important finding is that only 7.3\%  of the final solutions do not use \gls{dr} and that the most used \gls{dr} technique is \gls{is} alone (52.9\%), compared with only using \gls{fs} which represents only 17.6\% of the cases. This discovery supports the previous point that features are more important so we can reduce more at the instance level.  

Additionally, we explored whether using all the data available (train and validation from the CV) would lead to better results, as some studies stated \cite{Boonyanunta_2004_Predicting,L_2012_Predicting}.
In this case, we only evaluated it for the \gls{edca} and FLAML, as TPOT automatically retrains the pipeline using the entire training dataset at the instance level and does not permit retraining at the feature level.
The results only show differences in \gls{edca} when more data is added (see Table \ref{tab:mcc-values}). When the frameworks are retrained with the entire data, \gls{edca} surpasses the results achieved by the other two frameworks. In addition, the results reveal no differences between the FLAML with all the data and \gls{edca} using only the selected data. Therefore, in the light of \gls{greenautoml}, it is better to use \gls{edca} with less data since they achieve identical results. Comparing the FLAML with all training data or only with the selected data, it can be concluded that MCC did not improve when the results were rounded to two decimal points. 

To verify whether combining multiple \gls{automl} frameworks was helpful and whether the data chosen using \gls{edca} was most suitable for the problem, the best pipelines found with FLAML and TPOT retrained with the data optimised by \gls{edca}. Table \ref{tab:mcc-values-retrained} shows a performance loss when the best solutions on FLAML and TPOT are retrained with \gls{edca}'s data. This loss can be from various sources: (i) the frameworks used different preprocessing optimisations that may influence the results; (ii) the frameworks achieved different final classification models, and the ones achieved by FLAML and TPOT may require more data than the ones achieved with \gls{edca}. In short, optimising the entire pipeline with \gls{edca} achieved good results since the pipeline evolved with all the components, but testing only one part of it may not result in improvements.

\begin{figure}[t!]
    \centering
    \includegraphics[width=\linewidth]{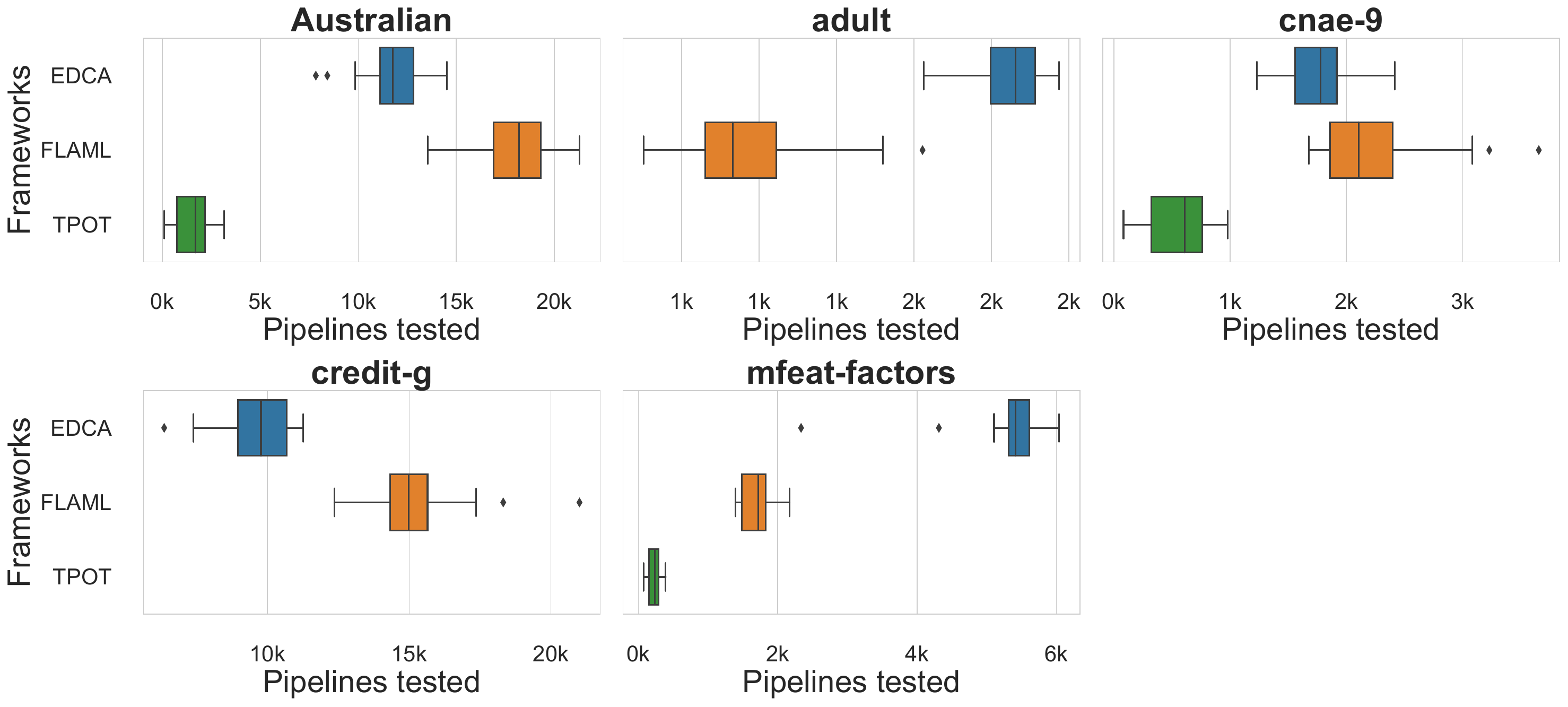}
    \caption{
    Number of pipelines evaluated by each framework over 30 runs.
    }
    \label{fig:boxplot-number-evaluations-frameworks}
\end{figure}

Another aspect analysed was how many solutions each framework was able to test.
Figure \ref{fig:boxplot-number-evaluations-frameworks} shows that the size of the search space of each framework is inversely related to the number of evaluations made. FLAML has a smaller search space since it only selects the most appropriate classification model and tests more solutions in almost all datasets. Then, \gls{edca} optimises the entire \gls{ml} pipeline in a linear format, being the second framework that tests more solutions. Finally, TPOT has a larger search space as it represents the solutions as graphs, which are more complex, evaluating, and therefore, fewer solutions. 
The frameworks that test more solutions, FLAML and \gls{edca}, also apply \gls{is} during the optimisation, which accelerates the optimisation. 

\begin{table}[t!]
    \centering
    \caption{Best pipelines achieved by EDCA, FLAML and TPOT across all runs and folds. "SI" is SimpleImputer. "\%I"  and "\%F" mean the instances and features percentage, respectively. "Imp" is Imputer.}
    \resizebox{\linewidth}{!}{
    \begin{tabular}{r|r|r|r|r|r|r|p{4cm}|r}
    \toprule
    \textbf{Dataset} & \textbf{Framework} & \textbf{\%I} & \textbf{\%F} & \textbf{Imp} & \textbf{Scaler} & \textbf{Encoder} & \textbf{Model} & \textbf{MCC} \\
    \midrule
    adult & EDCA & 0.63 & 1.0 & SI & Standard & One-Hot & XGBClassifier & 0.657\\
    adult & FLAML & 0.75 & 1.0 & - & - & - & LGBMClassifier & 0.658\\
    adult & TPOT & - & - & - & - & - & - \\
    \hline
    Australian & EDCA & 0.33 & 1.0 & - & MinMax & - & XGBClassifier & 0.853\\
    Australian & FLAML & 0.75 & 1.0 & - & - & - & XGBClassifier & 0.854\\
    Australian & TPOT & 1.0 & 0.5 & - & - & - & LGBMClassifier & 0.827\\
    \hline
    cnae-9 & EDCA & 0.75 & 1.0 & - & Standard & - & LogisticRegression & 0.974\\
    cnae-9 & FLAML & 0.75 & 1.0 & - & - & - & XGBClassifier & 0.969\\
    cnae-9 & TPOT & 1.0 & 1.0 & - & - & - & StackingEstimator (ExtraTreesClassifier), Standard, LogisticRegression  & 0.979\\
    \hline
    credit-g & EDCA & 0.5 & 1.0 & - & Robust & One-Hot & LogisticRegression & 0.467\\
    credit-g & FLAML & 0.75 & 1.0 & - & - & - & LGBMClassifier & 0.577\\
    credit-g & TPOT & - & - & - & - & - & - \\
    \hline
    mfeat-factors & EDCA & 0.75 & 0.78 & - & Standard & - & LogisticRegression & 0.983\\
    mfeat-factors & FLAML & 0.75 & 1.0 & - & - & - & XGBClassifier & 0.986\\
    mfeat-factors & TPOT & 1.0 & 1.0 & - & Standard & - & LogisticRegression & 0.989\\
    \bottomrule
    \end{tabular}
    }
    \label{tab:final-solutions-frameworks}
\end{table}

\begin{figure}[h]
    \centering
    \includegraphics[width=\linewidth]{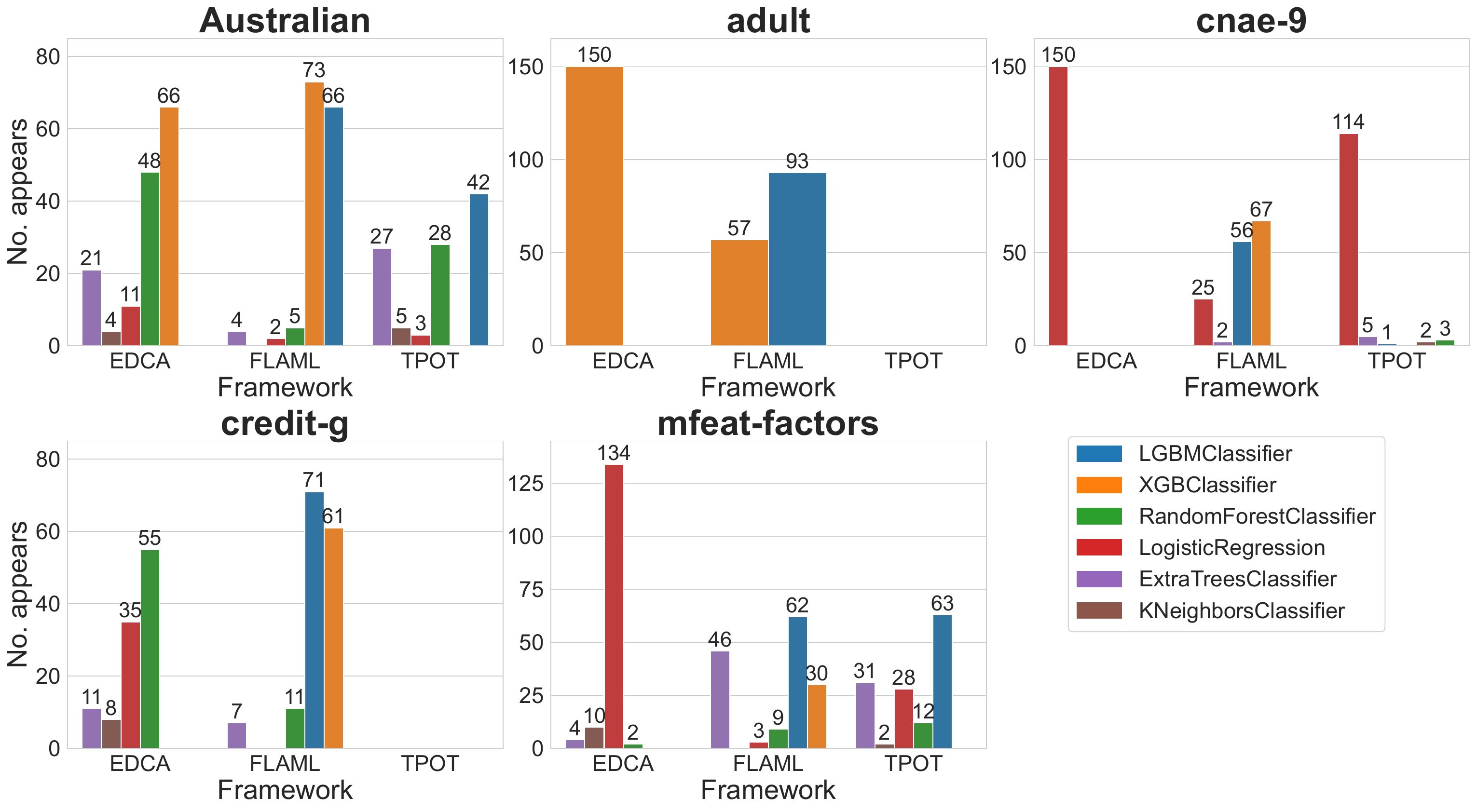}
    \caption{Distribution of the frameworks' best models in a 5-fold CV over 30 runs.}
    \label{fig:hist-models-frameworks}
\end{figure}

Additionally, we studied the final solutions each framework created (see Table \ref{tab:final-solutions-frameworks}) and concluded that there are differences and the type of solutions can also explain the differences considering the number of evaluations each one had made. 
\gls{edca} creates complete \gls{ml} pipelines and, in most cases, it was the only one who had preprocessing steps as predicted, despite TPOT also having those steps on their search space. Despite this, EDCA and TPOT varied more in the final solutions than FLAML.
Furthermore, no pattern was found in the distribution of the final models chosen by the frameworks, and they do not agree on the best one (see Figure \ref{fig:hist-models-frameworks}). Also, FLAML tends only to choose gradient-boosting algorithms. 
In addition, due to the representation used, TPOT applies ensemble learning in some cases, allowing for more complex solutions. 
Thus, based on the results in this set of experiments, \gls{edca} generates complete, simpler and efficient solutions trained with less data that, according to the statistical significance tests performed, have similar performance to state-of-the-art frameworks.


\section{Conclusion}
\label{sec:conclusion}

\gls{automl} has been useful to accelerate the development of \gls{ml} systems. However, despite knowing that data quality impacts the accuracy of the \gls{ml}, most \gls{automl} frameworks efforts only focus on predictive model tuning. Additionally, \gls{automl} is criticised for its resource-intense optimisation.
Therefore, in this paper, we proposed a data-centric \gls{automl} framework called \gls{edca} - Evolutionary Data Centric AutoML, focused on improving the entire \gls{ml} classification pipeline from data processing to model tuning, while also reducing the data to lower costs. \gls{edca} starts by analysing the given dataset to infer which data transformations must be applied. Considering the costs associated with training the \gls{ml} systems, it includes a component of automatic data optimisation to select the most relevant data (instances and features) for the training. Using \gls{dr} during the optimisation and in the final pipelines allows for the reduction of the associated costs. 
Like other \gls{automl} frameworks, \gls{edca} also selects the best predictive model for the problem and for this end, a \gls{ga} optimises the pipeline.

\gls{edca} was compared to state-of-the-art frameworks, FLAML and TPOT, on benchmark datasets. When evaluated on the same conditions, \gls{edca} achieved similar predictive performance results to others. The reduction and transformation of the received data did not put \gls{edca} at a disadvantage against FLAML and TPOT. \gls{edca} uses significantly less data on the final solutions, which in addition to the MCC values achieved by the three frameworks, shows that \gls{edca} is a low-cost \gls{automl} capable of creating simpler but efficient \gls{ml} solutions. This positive outcome indicates that it is possible to accomplish good results with low costs by optimising the data.
In summary, given the concerns around resource-intensive \gls{automl} frameworks, based on the results attained so far suggest that \gls{edca} offers a more efficient and cost-effective \gls{automl} solution, aligning with the principles of \gls{greenautoml}.

In the future, we will be conducting experiments about fitness function's impact on the final \gls{ml} pipelines and the impact of \gls{dr} on the evolutionary stage in terms of costs, while incorporating more \gls{greenautoml} guidelines. Also, we want to make a deeper study about the \gls{ea} parameters used, which were obtained through empirical testing. Finally, building on our initial results, we aim to extend the comparison with literature by testing \gls{edca} against other \gls{automl} frameworks. Furthermore, \gls{edca} will also be evaluated on more benchmark datasets and real-world datasets to see its generalisation capabilities.

\subsubsection{Acknowledgements}

This work has been partially supported by Project 
"NEXUS Pacto de Inovação – Transição Verde e Digital para Transportes, Logística e Mobilidade". ref. No. 7113, supported by the Recovery and Resilience Plan (PRR); by the Portuguese Recovery and Resilience Plan (PRR) through project C645008882-00000055, Center for Responsible AI; by the European Funds Next Generation EU, following Notice No. 02/C05-i01/2022.PC645112083-00000059 (project 53), Component 5 - Capitalization and Business Innovation - Mobilizing Agendas for Business Innovation; by the FCT - Fundação para a Ciência e a Tecnologia, I.P., in the framework of the Project UIDB/00326/2025 and UIDP/00326/2025.

\bibliographystyle{splncs04}
\bibliography{references}

\end{document}